\documentclass[journal]{IEEEtran}
\ifCLASSINFOpdf
   \usepackage[pdftex]{graphicx}
\else
\fi
\usepackage{amsmath}
\usepackage{algorithm,algpseudocode}

\usepackage[justification=centering]{caption}
\ifCLASSOPTIONcompsoc
  \usepackage[caption=false,font=normalsize,labelfont=sf,textfont=sf]{subfig}
\else
  \usepackage[caption=false,font=footnotesize]{subfig}
\fi
\usepackage{color}

\usepackage{stfloats}

\usepackage{cleveref}

\begin{document}
%
\title{Context-Aware Semantic Inpainting}
%
%
%

\author{Haofeng~Li,
        Guanbin~Li,
        Liang~Lin,
        and~Yizhou~Yu 
\thanks{H. Li and Y. Yu are with the Department
of Computer Science, The University of Hong Kong, HK}
\thanks{G. Li and L. Lin are with the school of Data and Computer Science,
Sun Yat-sen University, Guangzhou 510006, China.}
}

%
%

\markboth{Journal of \LaTeX\ Class Files,~Vol.~14, No.~8, August~2015}%
{Shell \MakeLowercase{\textit{et al.}}: Bare Demo of IEEEtran.cls for IEEE Journals}
%



\maketitle

\begin{abstract}
Recently image inpainting has witnessed rapid progress due to generative adversarial networks (GAN) that are able to synthesize realistic contents. However, most existing GAN-based methods for semantic inpainting apply an auto-encoder architecture with a fully connected layer, which cannot accurately maintain spatial information. In addition, the discriminator in existing GANs struggle to understand high-level semantics within the image context and yield semantically consistent content. Existing evaluation criteria are biased towards blurry results and cannot well characterize edge preservation and visual authenticity in the inpainting results. In this paper, we propose an improved generative adversarial network to overcome the aforementioned limitations. Our proposed GAN-based framework consists of a fully convolutional design for the generator which helps to better preserve spatial structures and a joint loss function with a revised perceptual loss to capture high-level semantics in the context. Furthermore, we also introduce two novel measures to better assess the quality of image inpainting results. Experimental results demonstrate that our method outperforms the state of the art under a wide range of criteria.
\end{abstract}


%
\IEEEpeerreviewmaketitle

\section{Introduction}\label{sec:intro}
%
%
%
%

\IEEEPARstart{I}{mage} inpainting aims at synthesizing the missing or damaged parts of an image. It is a fundamental problem in low-level vision and has attracted widespread interest in the computer vision and graphics communities as it can be used for filling occluded image regions or repairing damaged photos. Due to the inherent ambiguity of this problem and the complexity of natural images, synthesizing content with reasonable details for arbitrary natural images still remains a challenging task.

High-quality inpainted result should be not only realistic but also semantically consistent with the image context surrounding the missing or damaged region at different scales. First, colorization should be reasonable and spatially coherent. Second, structural features such as salient contours and edges should be connected inside the missing region or across its boundary. Third, texture generated within the missing region should be consistent with the image context and contains high-frequency details. In addition, missing object parts need to be recovered correctly, which is challenging and requires capturing high-level semantics.

\begin{figure}[!t]
\centering
\subfloat[Input]{\includegraphics[width=1.3in]{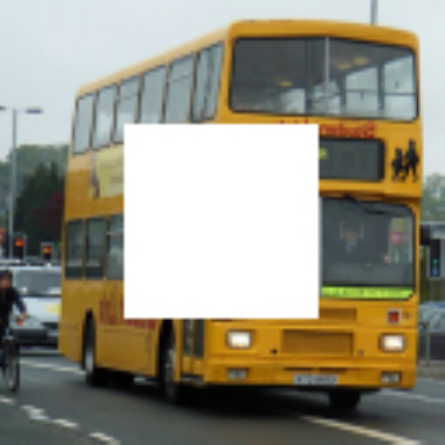}%
\label{fig_first_case}}
\hspace{0.05cm}
\subfloat[CASI]{\includegraphics[width=1.3in]{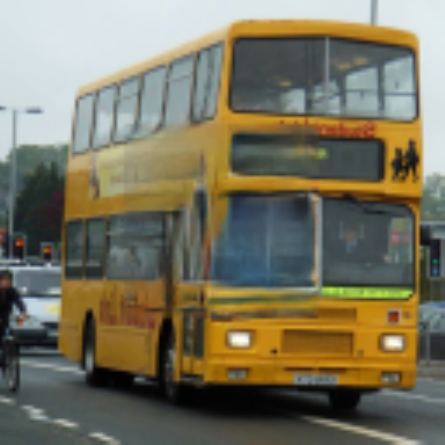}%
\label{fig_second_case}}
\vspace{0.2cm}
\subfloat[Conten-Aware Fill]{\includegraphics[width=1.3in]{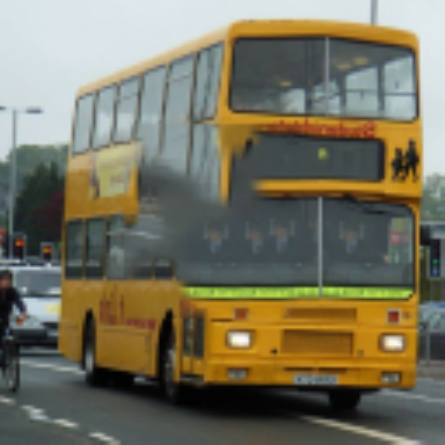}%
\label{fig_third_case}}
\hspace{0.05cm}
\subfloat[Context Encoder]{\includegraphics[width=1.3in]{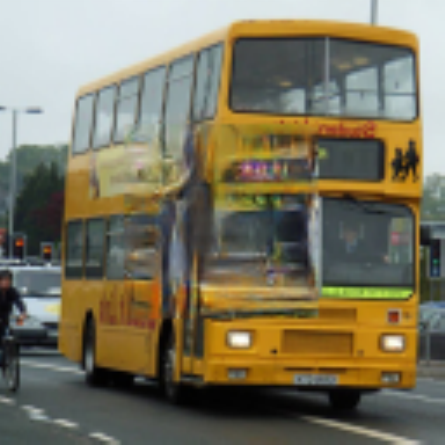}%
\label{fig_forth_case}}
\caption{Our proposed CASI with perceptual loss synthesizes content with a more reasonable colorization and structure than Content-Aware Fill \cite{contentAwareFill} and Context Encoder~\cite{pathak2016context}.}
\label{fig_sim}
\end{figure}

Deep convolutional neural networks are capable of learning powerful image representations and have been applied to inpainting \cite{NIPS2012_4686,Fawzi2016Image} with varying degrees of success. Recently semantic image inpainting has been formulated as an image generation problem and solved within the framework of generative adversarial networks (GAN)~\cite{goodfellow2014generative}. GAN trains a generator against a discriminator and successfully generates plausible visual content with sharp details. State-of-the-art results~\cite{pathak2016context,yeh2016semantic,yang2016high} have been achieved.

However, all existing GAN-based solutions to inpainting share common limitations. First of all, they utilize an encoder-decoder architecture with fully connected layers as the {\textit{bottleneck}} structure in the middle of the network. The bottleneck structure contains two fully connected (fc) layers. The first fc layer converts convolutional features with spatial dimensions to a single feature vector and another fc layer maps the feature vector backward to features with spatial information. The first fully connected layer collapses the spatial structure of the input image so that location related information cannot be accurately recovered during the decoding process. Second, the discriminator only takes a synthesized region without its image context as the input. Thus neither structural continuity nor texture consistency can be guaranteed between the synthesized region and its image context. Moveover, existing GANs struggle to understand high-level semantics within the image context and yield semantically consistent content.

To overcome the aforementioned limitations, we conceive a novel fully convolutional generative network for semantic inpainting. First, we adopt a fully convolutional design without the bottleneck structure to preserve more spatial information. Second, we composite the synthesized region and its image context together as a whole, and measures the similarity between this composite image and the ground truth. To increase such similarity, a perceptual loss is computed for the composite image. This perceptual loss defined in terms of high-level deep features is promising in capturing the semantics of the image context.

Furthermore, noticing that the $L2$ loss and PSNR are unable to rate blurry results accurately and quantitative measures do not exist for assessing how well the intended semantics have been restored, we define a local entropy error and a semantic error to resolve these two issues respectively. The semantic error (SME) is defined as the hinge loss for the confidence that a composite image with a synthesized region should be assigned the groundtruth label of its real counterpart, where the confidence value is estimated by a pre-trained image classifier. In our experiments, images synthesized by our inpainting model can successfully reduce the semantic error estimated by a powerful image classifier. This indicates that our model is capable of inferring semantically valid content from the image context.

In summary, this paper has the following contributions:
\begin{itemize}
\item We present a fully convolutional generative adversarial network without a fully-connected layer for maintaining the original spatial information in the input image. This network can process images with a variable size.

\item We introduce a novel context-aware loss function including a perceptual loss term, which measures the similarity between a composite image and its corresponding groundtruth real image.

\item We propose two novel measures, a local entropy error based on middle-level statistics and a semantic error based on high-level features, for evaluating inpainting results.
\end{itemize}


\section{Related Work}
Recently, deep neural networks including generative adversarial networks have exhibited great performance in image generation, image transformation and image completion. This section discusses previous work relevant to image inpainting and our proposed method.

\subsection{Image Inpainting}
Many algorithms on recovering holes in images or videos have been proposed \cite{huang2014image}, \cite{jia2003image}, \cite{rares2005edge}, \cite{iizuka2017globally}, \cite{pritch2009shift}, \cite{wexler2004space}, \cite{wexler2007space}, \cite{sun2005image}. Some existing methods for image completion are related to texture synthesis \cite{efros1999texture},~\cite{efros2001image} or patch-based synthesis \cite{criminisi2003object}, \cite{xu2010image}, \cite{darabi2012image}. Efros and Leung~\cite{efros1999texture} proposed a method for predicting pixels from the context boundary while \cite{efros2001image} searches for matching patches and quilts them properly. Drori et.al.~\cite{drori2003fragment} computed a confidence map to guide filling while Komodakis et.al.~\cite{komodakis2007image} proposed a priority belief propagation method. However, these exemplar based approaches struggle to generate globally consistent structures despite producing seamless high-frequency textures. Hays and Efros~\cite{Hays:2007} filled large missing regions using millions of photographs and presented seamless results. However, in this method, missing regions need to be prepared carefully by completely removing partially occluded objects. Synthesizing content for arbitrary missing regions remains a challenging task (e.g., recovering body parts for a partially occluded object).

\subsection{Generative Adversarial Networks}
Generative adversarial networks (GAN), which estimate generative models by simultaneously training two adversarial models were first introduced by Goodfellow et.al.~\cite{goodfellow2014generative} for image generation. Radford et.al.~\cite{radford2015unsupervised} further developed a more stable set of architectures for training generative adversarial networks, called deep convolutional generative adversarial networks~(DCGAN).
Recently GAN has widely applied to image generation \cite{dosovitskiy2016generating}, image transformation\cite{pix2pix2016}, image completion \cite{pathak2016context} and texture synthesis \cite{li2016precomputed}. Context Encoder~\cite{pathak2016context} uses a novel channel-wise fully connected layer for feature learning but keeps the traditional fully connected layer for semantic inpainting. Yeh et.al.~\cite{yeh2016semantic} employed GAN with both a perceptual loss and a contextual loss to solve inpainting. Notice that the perceptual loss in \cite{yeh2016semantic} is essentially an adversarial loss and the contextual loss considers the context only (excluding the synthesized region). Yang et.al.~\cite{yang2016high} conducted online optimization upon a pre-trained inpainting model primarily inherited from Context Encoder. The optimization is too expensive for real-time or interactive applications.
Common disadvantages exist in these GAN based approaches. First, the fully connected layer in the encoder-decoder framework cannot preserve accurate spatial information. Second, the discriminator in current GANs only evaluates the synthesized region but not the semantic and appearance consistency between the predicted region and the image context.

\subsection{Fully Convolutional Networks}
Fully convolutional networks (FCNs), which was first used in \cite{long2015fully} for semantic image segmentation, provides an end-to-end learnable neural network solution for pixel-level image comprehension. Without fully connected layers, FCNs occupy less memory and can learn and predict more efficiently. Besides, FCNs preserve spatial information and extract location sensitive features. Recently FCNs have achieved excellent results on semantic segmentation \cite{long2015fully}, edge detection \cite{Xie_2015_ICCV}, saliency detection \cite{LiYu16} and other pixel-wise labeling tasks. In this paper, we exploit the idea of FCN in GAN-based inpainting to better capture object contours, preserve spatial information in features, and infer coherent visual content from context.

\subsection{Context-Aware Perceptual Loss}
Perceptual loss is a feature reconstruction loss defined by deep neural networks~\cite{johnson2016perceptual}. It guides neural models to generate images visually similar to their corresponding targets~(e.g., ground truth) and has been widely utilized in style transfer~\cite{gatys2015neural}. Dosovitskiy and Brox et.al.~\cite{dosovitskiy2016generating} presented a similar concept, called DeePSiM, which successfully generates images with sharp details.
So far perceptual loss has been applied to style transfer~\cite{gatys2015neural,johnson2016perceptual}, super resolution~\cite{johnson2016perceptual} and texture synthesis\cite{ulyanov2016texture}. However, these topics primarily use the ``texture network'', a part of the VGG network~\cite{Simonyan14c} to extract middle-level features while high-level features from the fully connected layers have not been investigated for image completion. In this paper we exploit high-level deep features in the definition of perceptual loss to synthesize regions  semantically consistent with their contexts.

\section{Method}



\begin{figure*}[ht]
\centering

\subfloat[Fully Convolutional Generative Network]{\includegraphics[width=0.85\textwidth]{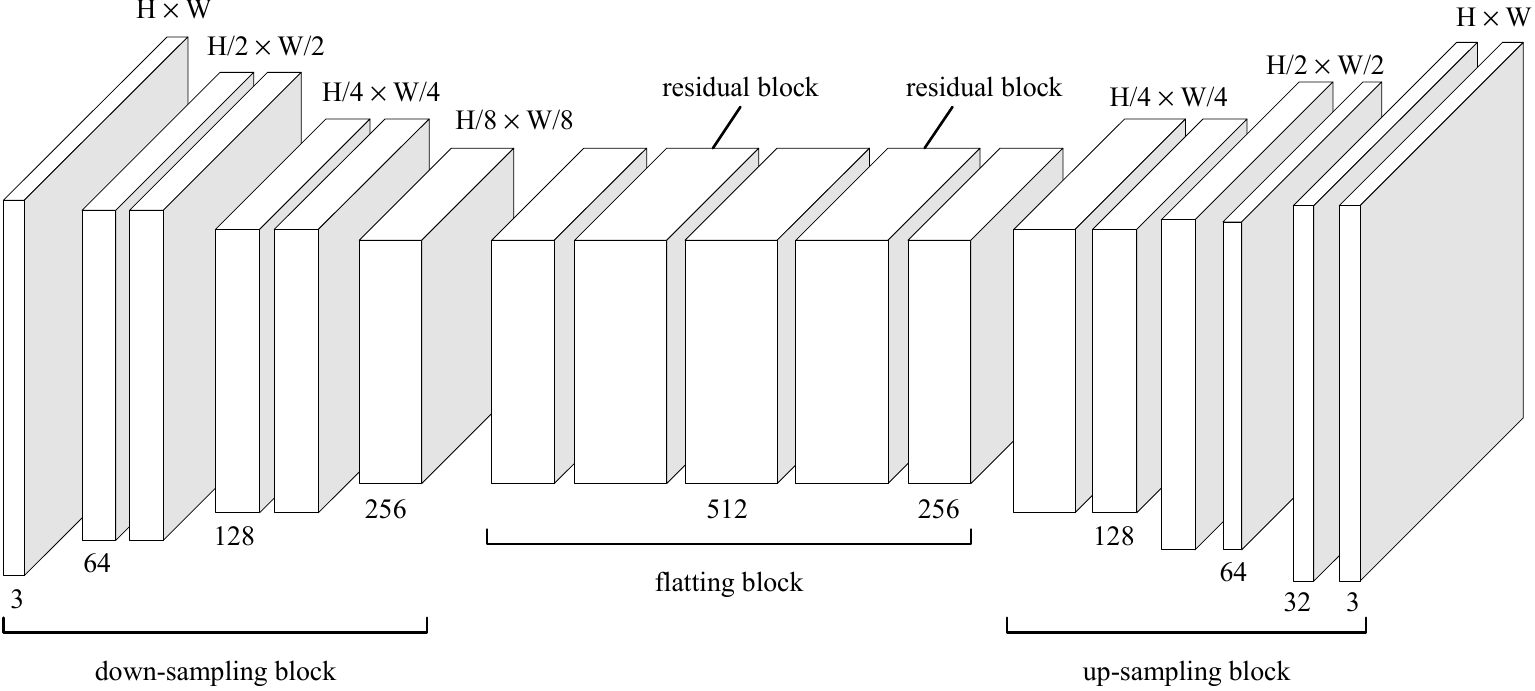}%
\label{fig:gen}}
\hfil
\subfloat[Context-Aware Semantic Inpainting Pipeline]{\includegraphics[width=0.85\textwidth]{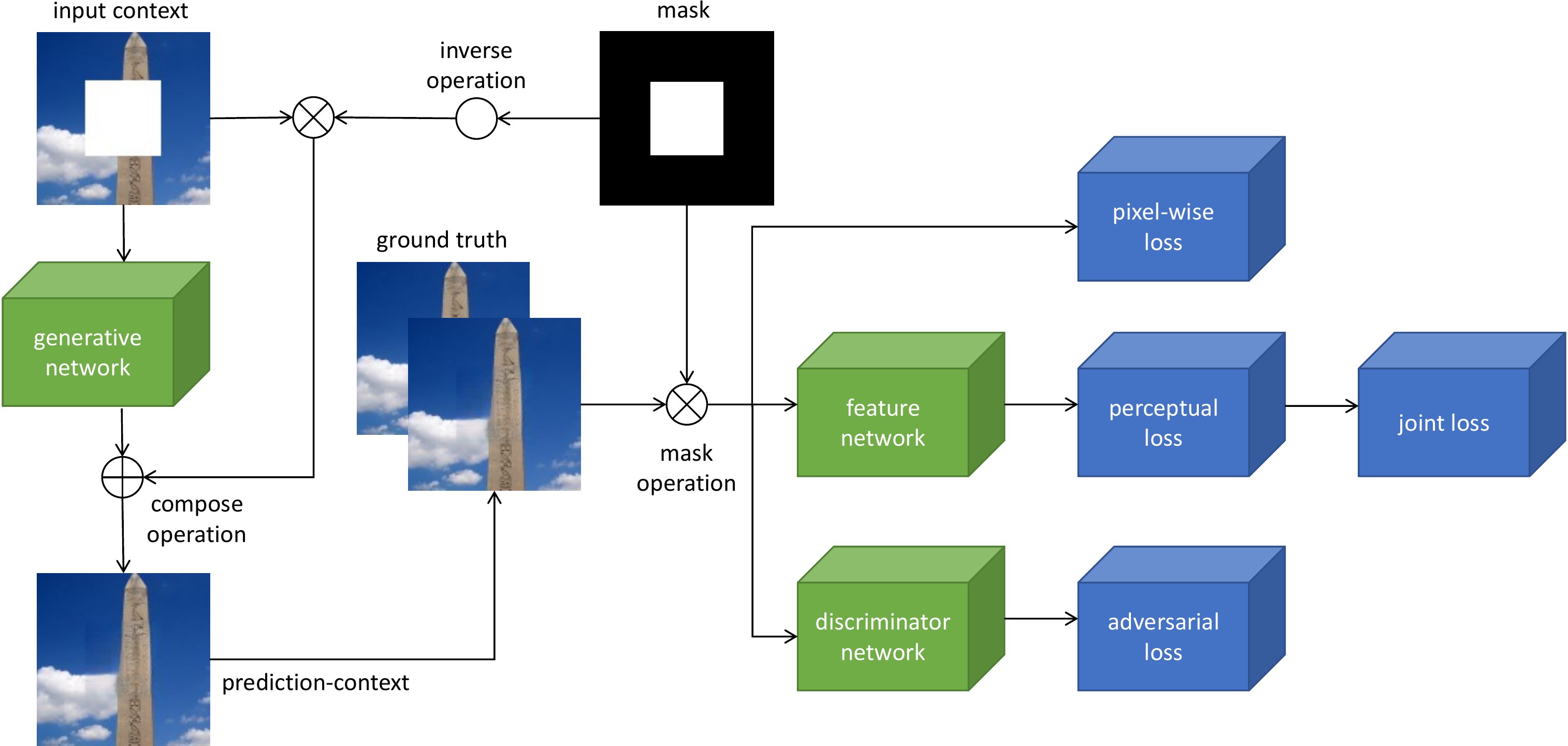}%
\label{fig:pipeline}}
\hfil
\caption{Network Architecture}
\label{fig:CASI}
\end{figure*}



As shown in Figure~\ref{fig:pipeline}, our proposed Context-Aware Semantic Inpainting method (CASI) is composed of an inpainting generation pipeline (on the left) and a joint loss function (on the right).
The fully convolutional generative network takes an image context as the input, where the missing region is filled with the mean pixel value. The missing region is generated by point-wise multiplication (denoted as `mask operation') with a mask. The inverse operation turns one into zero, and zero into one. The output of the generative network is a synthesized image with the same size as the input.  Then this output image is cropped using the boundary of the missing region and placed within the image context to form a composite image (denoted as `prediction-context'), via a point-wise addition (denoted as `compose operation'). The discriminator network receives the synthesized content within the missing region and the ground truth within the same region respectively, and attempts to classify the received content as either `real' or `fake'. The classification error is formulated as the adversarial loss, one of the components in the proposed loss. Our joint loss function is a linear combination of a pixel-wise $L2$ loss, the adversarial loss and a perceptual loss.

\subsection{Fully Convolutional Generative Network}
The fully convolutional generative network consists of three blocks: down-sampling, flatting and up-sampling. First, the down-sampling block plays the role of an encoder, which reduces each spatial dimension to 1/8 of the input size. The flatting block discovers and maintains essential edges without further changing the spatial size. Finally, the up-sampling block plays the role of a decoder, which transforms the feature map to an RGB image with the same resolution as the input.

The down-sampling block has three convolutional layers using 4$\times$4 kernels and two convolutional layers using 3$\times$3 kernels. The first layer of this block performs 4$\times$4 convolution. Then these two types of convolutional layers alternate and the block ends with a 4$\times$4 convolutional layer. The 4$\times$4 convolutions use a stride of 2 and 1 pixel padding to reduce the spatial size by half while doubling the number of channels in the feature map. Reduced spatial dimensions allow convolution kernels to have larger receptive fields in the input image. The 3$\times$3 convolutions use a stride of 1 and 1 pixel padding to keep the same spatial size and channel number. Such layers enhance the recognition capacity of the network. The flatting block has three convolutional layers using 3$\times$3 kernels and two residual blocks. These residual blocks enhance prediction accuracy for semantic inpainting. The middle layer doubles the number of channels while the last layer reduces it by half. Thus the flatting block keeps the number of channels the same in the input and output feature maps. The up-sampling block has three de-convolutional layers using 4$\times$4 kernels and three convolutional layers using 3$\times$3 kernels. Similar to the down-sampling block, the two types of layers alternate, and the first layer performs 4$\times$4 deconvolution. In the up-sampling block, 4$\times$4 deconvolution acts as parameterized interpolation which doubles the spatial size while each 3$\times$3 convolutional layer reduces the number of channels by half. The last layer of the up-sampling block generates an RGB image with the same size as the input.

Our proposed generative network does not have a bottleneck fully connected layer, and enjoys the benefits of fully convolutional architecture. It is capable of locating essential boundaries, maintaining fine details and yield consistent structures in missing regions.

\subsection{Discriminative Network}
Our discriminator shares a similar but shallower structure with the down-sampling block in the generator network. Compared with the down-sampling block, the discriminator removes all 3$\times$3 convolutional layers to avoid overfitting. Otherwise, the capacity of the discriminator would be so large that the generator does not have a chance to confuse the discriminator and improve itself. A fully connected layer is employed to perform binary classification at the end of the discriminator.

Normalization and non-linear activations are used in CASI. Except for the last layer, every convolutional layer in the generator and the discriminator is followed with a batch normalization (batchnorm) layer. Rectified linear units (ReLU) follow each batchnorm layer in the generator while Leaky-rectified Linear Units (LeakyReLU) are used in the discriminator according to the architecture guidelines in DCGAN. A Sigmoid layer is adopted in the last layer of the generator and the discriminator to map pixel and confidence values respectively.

\subsection{Loss Function}
Given the analysis in Section~\ref{sec:intro}, existing GAN based semantic inpainting methods fail to grasp
high-level semantics and synthesize semantically consistent content for the missing region. In this paper, we propose to composite the synthesized region and its image context together as a whole, and measures
the visual similarity between this composite image and the ground truth using a perceptual loss.
Our overall loss function consists of a pixel-wise $L2$ loss, an adversarial loss and a perceptual loss. It can be formulated as follows,
\begin{equation} \label{eq3}
L_{inp} = \lambda_{pix} l_{pix} + \lambda_{adv} l_{adv} + \lambda_{per} l_{per},
\end{equation}
where $L_{inp}$ denotes the overall inpainting loss. $l_{per}$, $l_{adv}$, $l_{pix}$ denote our perceptual loss, adversarial loss and pixel-wise $L2$ loss respectively while $\lambda_{per}$, $\lambda_{adv}$ and $\lambda_{pix}$ are the weights of the respective loss terms.

Pixel-wise $L2$ loss, $l_{pix}$, is a straightforward and widely used loss in image generation. It measures the pixel-wise differences between the synthesized region and its corresponding ground truth. $l_{pix}$ is defined in Eq. (\ref{pixel-wise loss}),
\begin{equation} \label{pixel-wise loss}
l_{pix}(x, z) = {||M \odot (x - z)||_2}^2,
\end{equation}
where $M$ is a binary mask where a value of $1$ indicates the missing region and a value of $0$ indicates the known context region, $\odot$ is the element-wise product, $x$ is the ground-truth image and $z$ is the corresponding inpainting result computed as in Eq. (\ref{z}),
\begin{equation} \label{z}
z = ( (1-M) \odot x ) \oplus (M \odot G((1-M) \odot x)),
\end{equation}
where $\oplus$ is the element-wise addition, G is the CASI generator, $(1-M) \odot x$ is the context region of $x$, and $M \odot G(\cdot)$ is the missing region in the generator's output. $\oplus$ in Eq. (\ref{z}) merges the known context region and the synthesized missing region to obtain the final inpainting result.

However, calculating loss within the image space cannot guarantee to generate an image perceptually similar to the ground truth as neural networks tend to predict pixel values close to the mean of the training data. In practice, the pixel-wise $L2$ loss only produces blurred images without clear edges or detailed textures. Thus we exploit an adversarial loss and a novel perceptual loss to overcome this problem.

The adversarial loss $l_{adv}$ is defined on the objective function of the discriminator. As the discriminator aims at distinguishing synthesized content from its corresponding ground truth, its objective is to minimize a binary categorical entropy $e$ in Eq. (\ref{adv loss}). 
\begin{equation} \label{adv loss}
\begin{split}
& e(D(M \odot x), D(M \odot z))\\
&= -[log(D(M \odot x)) + log(1 - D(M \odot z))],
\end{split}
\end{equation}
where $e$ denotes binary categorical entropy and $D$ is the CASI discriminator. The discriminator $D$ predicts the probability that the input image is a real image rather than a synthesized one. If the binary categorical entropy is smaller, the accuracy of the discriminator is better. Note that $D$ is not a pre-trained or constant model during the training stage. Instead, $G$ and $D$ are trained alternatively. As minimizing the binary categorical entropy $e$ is equivalent to maximizing the negative of the binary categorical entropy, the final objective value of the discriminator is described in the right side of Eq. (\ref{full adv loss}). As the generator acts as an adversarial model of the discriminator, it tends to minimize the negative of the binary categorical entropy. Thus the adversarial loss of the generator $l_{adv}$ can be formally described as
\begin{equation} \label{full adv loss}
l_{adv} = \max_D [log(D(M \odot x)) + log(1 - D(M \odot z))].
\end{equation}

$l_{adv}$ makes the synthesized region deviate from the overly smooth result obtained using the pixel-wise $L2$ loss as real images are not very smooth and typically have fine details. Although the adversarial loss promotes fine details in the synthesized result, it also has disadvantages. First, existing discriminators are unaware of the image context and do not explicitly consider the composite image consisting of both the synthesized region and the image context. Second, binary classification is not challenging enough for the discriminator to learn the appearance of different objects and parts. Note that semantic inpainting needs to not only synthesize textures consistent with the context but also recover missing object parts, which requires high-level features extracted from the image context. Thus we propose a perceptual loss based on high-level semantic features.

Our perceptual loss, $l_{per}$, is defined in Eq. (\ref{eq4}),
\begin{equation} \label{eq4}
\begin{split}
l_{per}(x, z) &= e(F(x), F(z)) \\
&= \frac{1}{C_j H_j W_j} {||F_j(x) - F_j(z)||_2}^2,
\end{split}
\end{equation}
where $F$ is a pre-trained feature network that extracts a generic global feature from the input, $F_j$ denotes the activations of the $j$-th layer of $F$, $F_j(x)$ and $F_j(z)$ are a $C_j \times H_j \times W_j$ tensor respectively. In our experiments, we use ResNet-18 pre-trained over the ImageNet dataset~\cite{russakovsky2015imagenet} as the feature network $F$, and the 512-dimensional feature from the second last layer of ResNet-18 as $F_j$. Similar high-level features extracted by $F$ give rise to similar generated images, as suggested in \cite{dosovitskiy2016generating}. In addition, a perceptual loss based on high-level features makes up for the missing global information typically represented in a fully connected layer in the generator. Different from DeepSiM, our feature is extracted from the composite image consisting of the synthesized region and the image context rather than from the synthesized region alone.

\section{Implementation}
Let us discuss the details of our inpainting pipeline. Training images for CASI require no labels. As shown in Algorithm~\ref{algo1}, the training stage consists of a limited number of iterations. During each training iteration, the discriminator is updated $Diters$ times and the generator is trained once. In each iteration that updates the discriminator, each training image is separated into an image center and an image context. The image center has the same size of the central region, and the image context is the image filled with the mean pixel value in the central region. The image center and image context of a training image form a training pair. The generator takes the image context as the input and synthesizes the image center. The discriminator attempts to distinguish the synthesized content from the ground-truth image center. The adversarial loss is calculated and then the parameters of the discriminator are updated. In the rest of each training iteration, the pixel-wise $L2$ loss is computed, the feature network extracts a feature from the composite image, and three loss functions are combined to obtain the joint inpainting loss. The generator is finally updated according to the joint loss. This process is repeated until the joint loss converges. In the testing stage, each testing image is first filled with the mean pixel value in the center and then passed to the CASI generator. The central region of the generator's output is cropped and pasted back into the testing image to yield the final inpainting result.

Our CASI is implemented on top of DCGAN~\cite{radford2015unsupervised} and Context Encoder~\cite{pathak2016context} in Torch and Caffe~\cite{jia2014caffe}. ADAM~\cite{kingma2014adam} is adopted to perform stochastic gradient descent. As in \cite{pathak2016context}, CASI predicts a larger region which overlaps with the context region (by 4px). $10\times$ weight is used for the pixel-wise $L2$ loss in the overlapping area. Using a TITAN XP GPU, training on a dataset of 20000 images  costs 3 to 4 days. Inpainting a single image takes less than 0.2 seconds. Recovering a batch of 20 images costs less than 1 second.

\begin{algorithm}
\caption{ }
\label{algo1}
\begin{algorithmic}[1]
\State $F \gets \Call{loadModel()}{}$
\State $G \gets \Call{initWeight()}{}, D \gets \Call{initWeight()}{}$
\For {$i \gets 1, maxIterations$}
\State $x, z, M$
\For {$j \gets 1, Diters$}
\State $x \gets \Call{sampleBatch()}{}$
\State Compute $z$ using Eq. (\ref{z})
\State Compute $l_{adv}$ using Eq. (\ref{adv loss})
\State Update $D$
\EndFor
\State $l_{pix} \gets \Call{MSE}{x,z}$
\State $f_x \gets F(x), f_z \gets F(z)$
\State Compute $l_{per}$ using Eq. (\ref{eq4})
\State Compute $L_{inp}$ using Eq. (\ref{eq3})
\State Update $G$
\EndFor
\end{algorithmic}
\end{algorithm}

\section{Evaluation}
This section evaluates our proposed deep neural network architecture and joint loss function on a subset of ImageNet~\cite{russakovsky2015imagenet} and the Paris StreetView dataset~\cite{pathak2016context,doersch2012what}. This subset contains 20 randomly sampled categories, denoted as ``ImageNet-20''. ImageNet-20 consists of $25,000$ training images and $1,000$ testing images. Paris StreetView contains $14,900$ training samples and 100 testing samples.

\subsection{Effectiveness of Perceptual Loss}
We first verify whether adding a perceptual loss improves the results. CASI is trained using 4 different loss functions respectively to compare their performance. For these loss functions, the hyper-parameters of CASI are set in the same way, and the perceptual loss is defined using the same feature extracted using the same feature network. The four loss functions are: (a) pixel-wise $L2$ loss, (b) $L2$ loss + perceptual loss, (c) $L2$ loss + adversarial loss, (d) $L2$ loss + adversarial loss + perceptual loss. In the following we use (a)-(d) to refer to these loss functions.

\begin{figure*}[!t]
\centering
\includegraphics[width=1.0\textwidth]{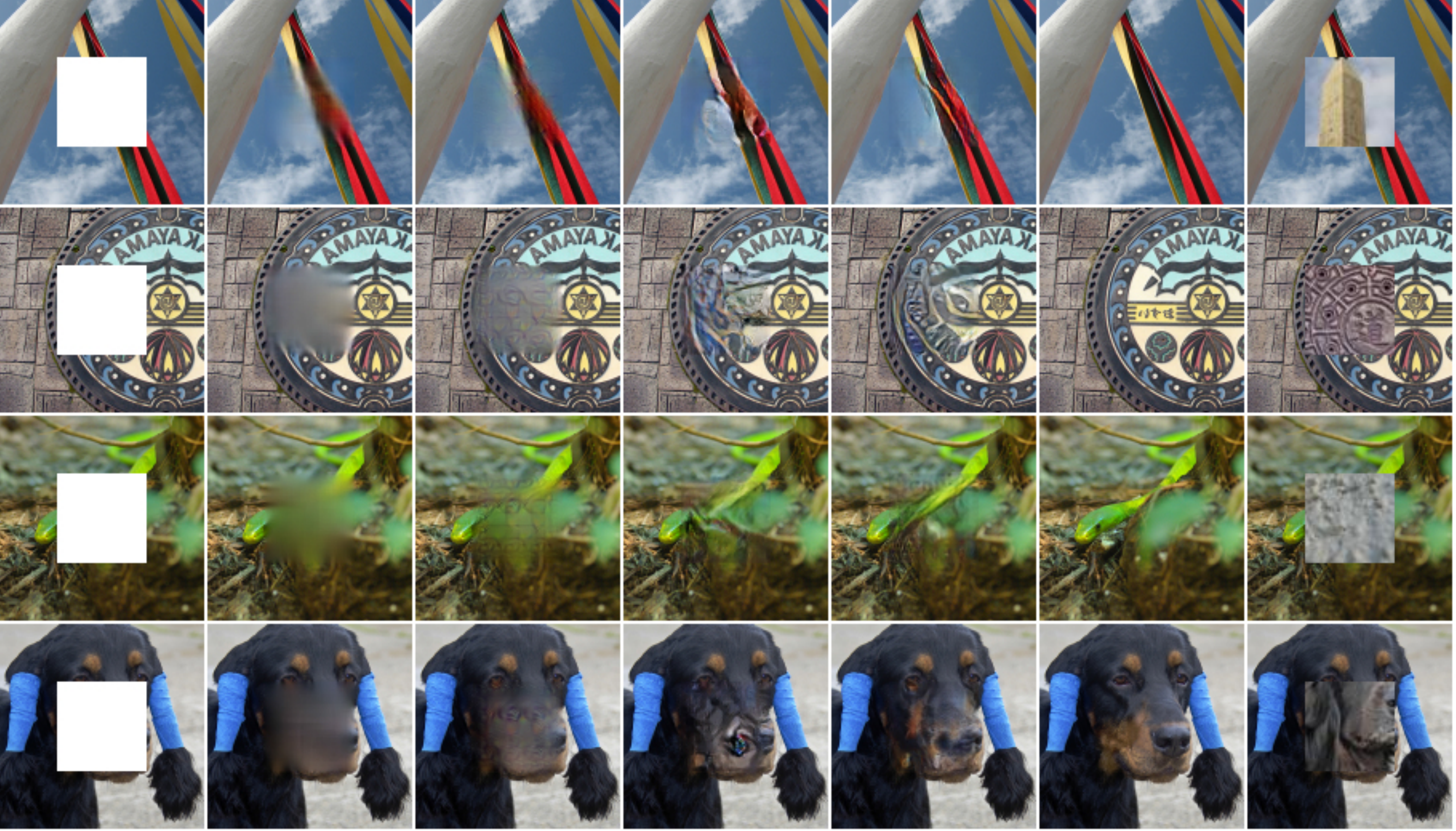}
\leftline{ \small \hspace{0.25cm} Input Image \hspace{1.45cm} (a) \hspace{1.97cm} (b) \hspace{1.97cm} (c) \hspace{1.97cm} (d) \hspace{1.35cm} Ground-Truth \hspace{0.6cm} NN-inpainting}
\leftline{ \small \hspace{3.4cm} $L2$ \hspace{1.7cm} $L2$+$per$ \hspace{1.25cm} $L2$+$adv$ \hspace{1.0cm} $L2$+$adv$+$per$}
\caption{ Comparison among different combinations of loss functions and Nearest-Neighbor(NN)-inpainting. The adversarial loss promotes low-level sharp details while the perceptual loss improves high-level semantic consistency.} 
\label{fig:internal}
\end{figure*}

Figure~\ref{fig:internal} show qualitative results of the above loss functions. The resolution of each images is $128 \times 128$. This result includes 4 samples representing different cases. All the missing regions are at the center of the image. From left to right, each column corresponds to a loss function from (a) to (d), respectively. As shown in this figure, (a) and (b) generate over-smooth results while (c) and (d) present sharper details. This conforms that the adversarial loss indeed alleviate the blurriness caused by the $L2$ loss. Between (a) and (b), (a) is more blurry while subtle textures or wrinkles can be observed in (b). Between (c) and (d), although they both preserve sharp edges, (d) is more semantically consistent with the context region. These results reveal that the adversarial loss works in the middle level to yield patches with consistent sharp details while the perceptual loss synthesizes consistent high-level contents.

\begin{table}
\renewcommand{\arraystretch}{1.3}
\caption{Quantitative results on ImageNet-20. CASIs without the adversarial loss achieve lower mean $L2$ error and higher PSNR but generate blurry results, which indicates that mean $L2$ error and PSNR inaccurately assess over-smooth cases.}
\label{table:quan1}
\small
\centering
\begin{tabular}{l|c|c|c}
\hline
Method                      & mean $L1$ & mean $L2$ & PSNR\\
                            & error     & error     &\\
\hline\hline
Context Encoder             & 12.15\%   &  3.31\%   &  15.59dB \\
CASI,$L2$                   & 11.07\%   &  \textbf{2.57\%}   &  \textbf{17.08dB} \\
CASI,$L2$ + $per$           & 11.21\%   &  2.64\%   &  16.95dB\\
CASI,$L2$ + $adv$           & 11.15\%   &  2.93\%   &  16.68dB \\
CASI,$L2$+$adv$+$per$       & \textbf{10.89\%}    &  2.83\%   &  16.81dB \\
\hline
\end{tabular}
\end{table}

Table~\ref{table:quan1} shows quantitative results from this experiment. It presents numerical errors between synthesized contents and their ground truth using three commonly employed measures, mean $L1$ error, mean $L2$ error and PSNR. Notations (a)-(d) are used to denote four trained CASI models. As shown in Table~\ref{table:quan1}, (a) achieves the smallest mean $L2$ error and PSNR while (d) achieves the smallest mean $L1$ error. Mean $L2$ error is smaller for solutions close to the mean value but such solutions are overly smooth and undesirable (see (a) and (b) in Figure~\ref{fig:internal}). Models trained without the adversarial loss have advantage in mean $L2$ error due to their blurry results. Similar results have been reported in \cite{dosovitskiy2016generating}. Between (c) and (d), (d) has smaller mean $L2$ error than (c). And (d) also has smaller mean $L1$ error than (c). Thus the perceptual loss is effective in improving our CASI model.

\subsection{Investigation of Perceptual Loss}
This section investigates how the parameter of perceptual loss effect the performance of our method. We set the hyper-parameters in our algorithm as follows. The summation of the weights of all loss terms is 1.0. The weight of the adversarial loss is 0.001, as suggested by \cite{pathak2016context}. We determine the weight of the perceptual loss $\lambda_{per}$ by cross validation on the ImageNet-20 dataset. As shown in Table~\ref{table:per}, setting the weight of the perceptual loss to 0.2 achieves the lowest mean L1 error, mean L2 error and the highest PSNR value among four different parameter settings.
\begin{table}
\renewcommand{\arraystretch}{1.3}
\caption{Investigation of Perceptual Loss}
\label{table:per}
\small
\centering
\begin{tabular}{l|c|c|c}
\hline
Method & mean $L1$ error & mean $L2$ error & PSNR\\
\hline \hline
$\lambda_{per}=0$ & 11.15\% & 2.93\% & 16.68dB\\
$\lambda_{per}=0.2$ & \textbf{10.89\%} & \textbf{2.83\%} & \textbf{16.81dB}\\
$\lambda_{per}=0.4$ & 11.12\% & 2.93\% & 16.60dB\\
$\lambda_{per}=0.7$ & 11.43\% & 3.06\% & 16.44dB\\
\hline
\end{tabular}
\end{table}

\begin{table}
\renewcommand{\arraystretch}{1.3}
\caption{ Effectiveness of Fully Convolutional Architecture }
\label{table:fc}
\small
\centering
\begin{tabular}{l|c|c|c}
\hline
Method & mean $L1$ error & mean $L2$ error & PSNR\\
\hline \hline
CASI+$fc$ & 9.70\% & 1.71\% & 18.83dB \\
CASI      & \textbf{7.49\%}  & \textbf{1.37\%} & \textbf{20.37dB} \\
\hline
\end{tabular}
\end{table}

 \subsection{Effectiveness of Fully Convolutional Architecture}
This section investigates whether applying fully convolutional architecture benefits semantic inpainting. We design a CASI+$fc$ model by inserting two fully connected layers after the third layer of the CASI flatting block~(described in Figure~\ref{fig:gen}). The first fully connected layer takes a convolutional feature map as input and outputs a 2048-d feature vector which is followed by a Tanh layer. The second fully connected layer takes the output of the activation layer as input and output a feature map with spatial dimensions. Then the fourth layer of the CASI flatting block takes the feature map as input. We compared CASI+$fc$ model and CASI model on Paris Street View dataset. As Table~\ref{table:fc} shows, CASI outperforms CASI+$fc$ by 2.21\% in mean $L1$ error, 0.34\% in mean $L2$ error and 1.54dB with regards to PSNR although CASI+$fc$ contains more parameters than CASI. The result suggests applying fully convolutional architecture is more conducive for generative network as the fully connected layers could collapse the spatial structure of the image features.

\subsection{Effectiveness of Residual Block}
This section verifies whether adding residual blocks enhance the performance. We design a CASI- model by removing the two residual blocks in CASI model and demonstrate comparison results between them. As shown in the upper part in Table~\ref{table:res}, CASI outperforms CASI- by 0.2\% in mean $L1$ error, 0.1\% in mean $L2$ error and 0.5dB in PSNR, on the ImageNet-20 dataset. As the lower part in Table~\ref{table:res} shows, CASI presents better performance than CASI- in mean $L1$ error, mean $L2$ error and PSNR value, on the Paris Street View dataset. The above results suggest that adding residual blocks improves prediction accuracy for the CASI model.

\begin{table}
\renewcommand{\arraystretch}{1.3}
\caption{Effectiveness of Residual Block}
\label{table:res}
\small
\centering
\begin{tabular}{l|c|c|c}
\hline
Method & mean $L1$ error & mean $L2$ error & PSNR\\
\hline \hline
CASI- & 11.09\%& 2.93\%& 16.31dB\\
CASI  & \textbf{10.89\%}& \textbf{2.83\%}& \textbf{16.81dB}\\
\hline \hline
CASI- & 7.79\%& 1.43\%& 20.14dB\\
CASI  & \textbf{7.49\%}& \textbf{1.37\%}& \textbf{20.37dB}\\
\hline
\end{tabular}
\end{table}

\subsection{High-resolution Case}
\begin{figure*}[t]
\centering
\includegraphics[width=0.8\linewidth]{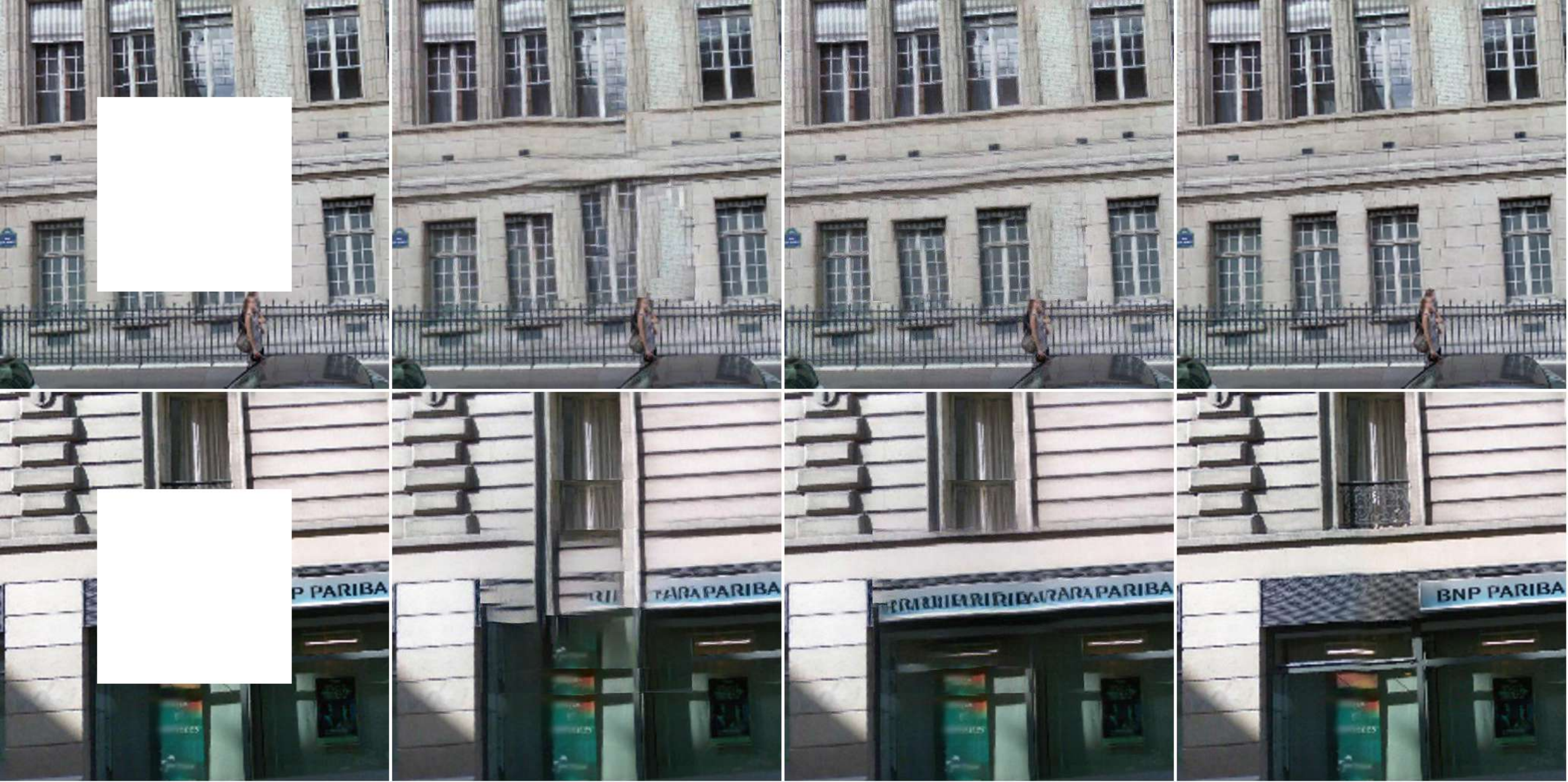}
\leftline{ \small \hspace{2.5cm} Input Image \hspace{2.0cm} NeuralPatch \hspace{2.0cm} CASI+ \hspace{2.0cm} Ground-Truth}
\caption{ High-resolution Cases on Paris StreetView} 
\label{fig:high-res-paris}
\end{figure*}

This section investigates how our method performs on high-resolution cases. The motivation of investigation on high-resolution cases is that most existing neural network based inpainting methods can only deal with input images not larger than $128 \times 128$. This section demonstrates how the proposed method perform with input images of $512 \times 512$. Two groups of experiments are presented. The first group compare our method to \cite{pathak2016context} by scaling image to match with the input size of \cite{pathak2016context}. As shown in upper part of Table~\ref{table:high}, our CASI model presents lower mean $L1$ error, lower mean $L2$ error and higher PSNR value than ContextEncoder~\cite{pathak2016context} in high-resolution Paris Street View dataset. The second group investigates whether adding a post-optimization based on our model deals with high-resolution cases. One concurrent work, NeuralPatch \cite{yang2016high}, trains its network to synthesize content at the image center and presents high-resolution object removal results during testing. We have integrated our method with post-optimization in~\cite{yang2016high} (denoted as CASI+) and demonstrate better performance than NeuralPatch~\cite{yang2016high}. As the lower part in Table~\ref{table:high} shows, the CASI+ method achieves lower mean $L1$ error, lower mean $L2$ error and higher PSNR value in comparison to NeuralPatch~\cite{yang2016high}, which suggests that the proposed CASI can provide more accurate reference content for post-optimization based image completion methods. Figure~\ref{fig:high-res-paris} is a qualitative comparison between \cite{yang2016high} and CASI+. As Figure~\ref{fig:high-res-paris} shows, CASI+ extends more reasonable edges and preserves more details than \cite{yang2016high}. More comparison results can be found in the supplementary document.
\begin{table}
\renewcommand{\arraystretch}{1.3}
\caption{High-resolution Case on Paris StreetView}
\label{table:high}
\small
\centering
\begin{tabular}{l|c|c|c}
\hline
Method & mean $L1$ error & mean $L2$ error & PSNR\\
\hline \hline
ContextEncoder& 9.04\% & 1.82\% & 18.90dB \\
CASI& \textbf{8.04\%} & \textbf{1.53\%} & \textbf{19.79dB} \\
\hline \hline
NeuralPatch& 9.59\% & 2.07\% & 18.42dB \\
CASI+& \textbf{8.62\%} & \textbf{1.73\%} & \textbf{19.18dB} \\
\hline
\end{tabular}
\end{table}

\begin{figure*}
\centering
\includegraphics[width=1.0\textwidth]{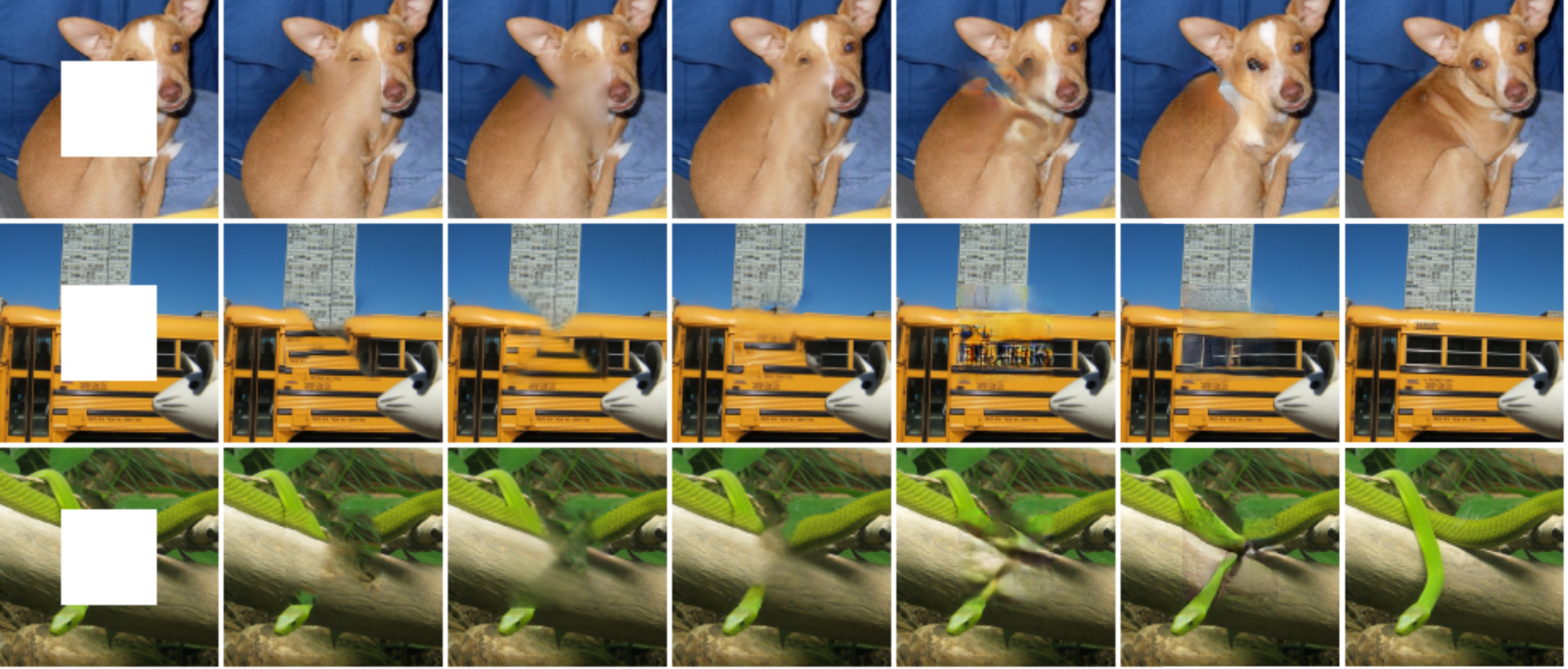}
\leftline{ \small \hspace{0.3cm} Input Image \hspace{0.28cm} Content-Aware Fill \hspace{0.05cm} StructCompletion
\hspace{0.35cm} ImageMelding \hspace{0.3cm} Context Encoder \hspace{0.95cm} CASI \hspace{1.1cm} Ground-Truth}
\caption{ Comparison on ImageNet-20 dataset }
\label{fig:more}
\end{figure*}

\begin{figure*}[t]
\centering
\includegraphics[clip, trim=0cm 14.25cm 0cm 0cm,
width=0.9\linewidth]{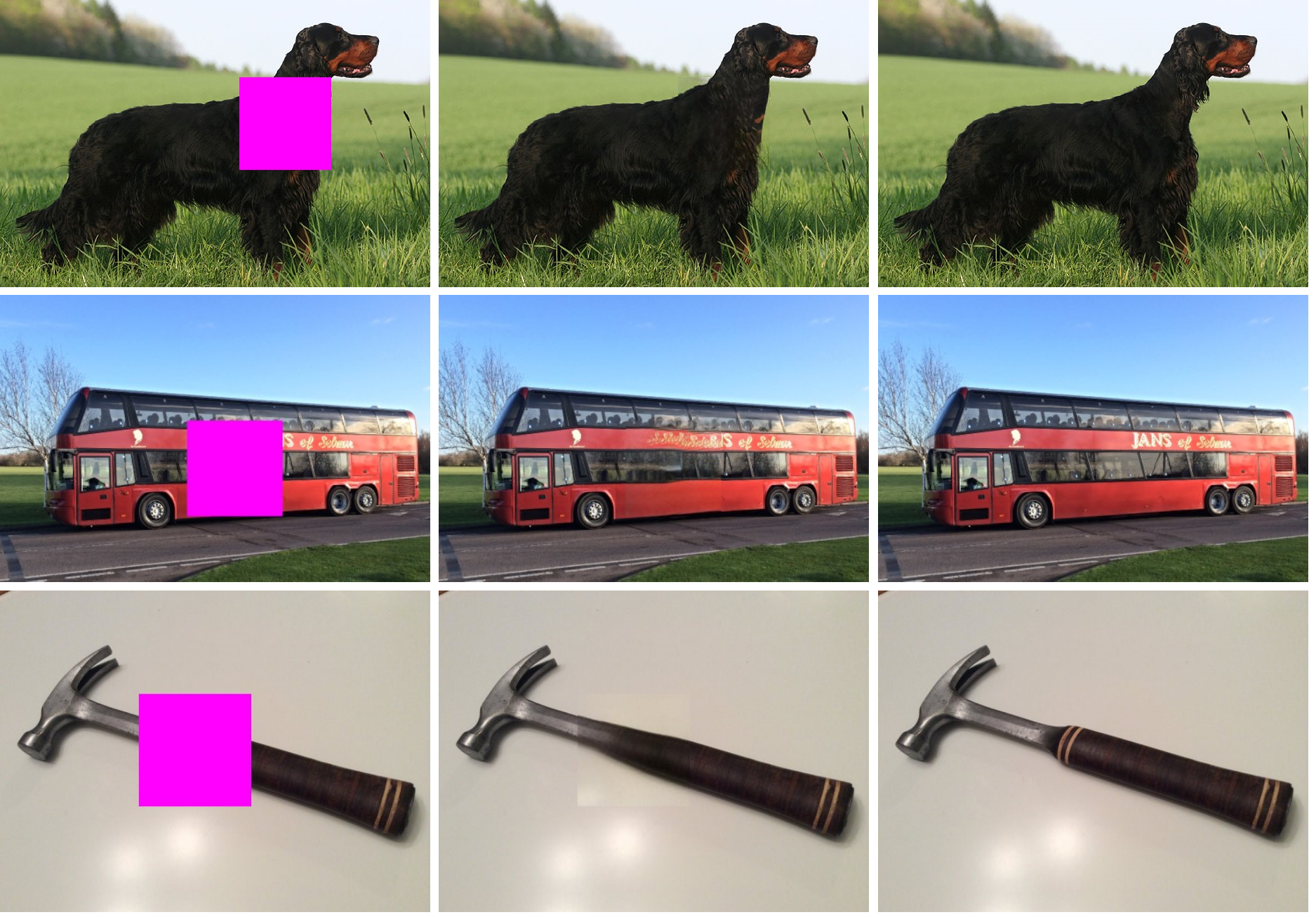}
\leftline{ \small \hspace{2.5cm} Input Image \hspace{4.0cm} CASI+ \hspace{4.0cm} Ground-Truth }
\caption{ Inpainting Results of CASI with In-the-wild Cases }
\label{fig:real-images}
\end{figure*}

\begin{figure*}[t]
\centering
\includegraphics[clip, trim=0cm 14.25cm 0cm 0cm,
width=0.9\linewidth]{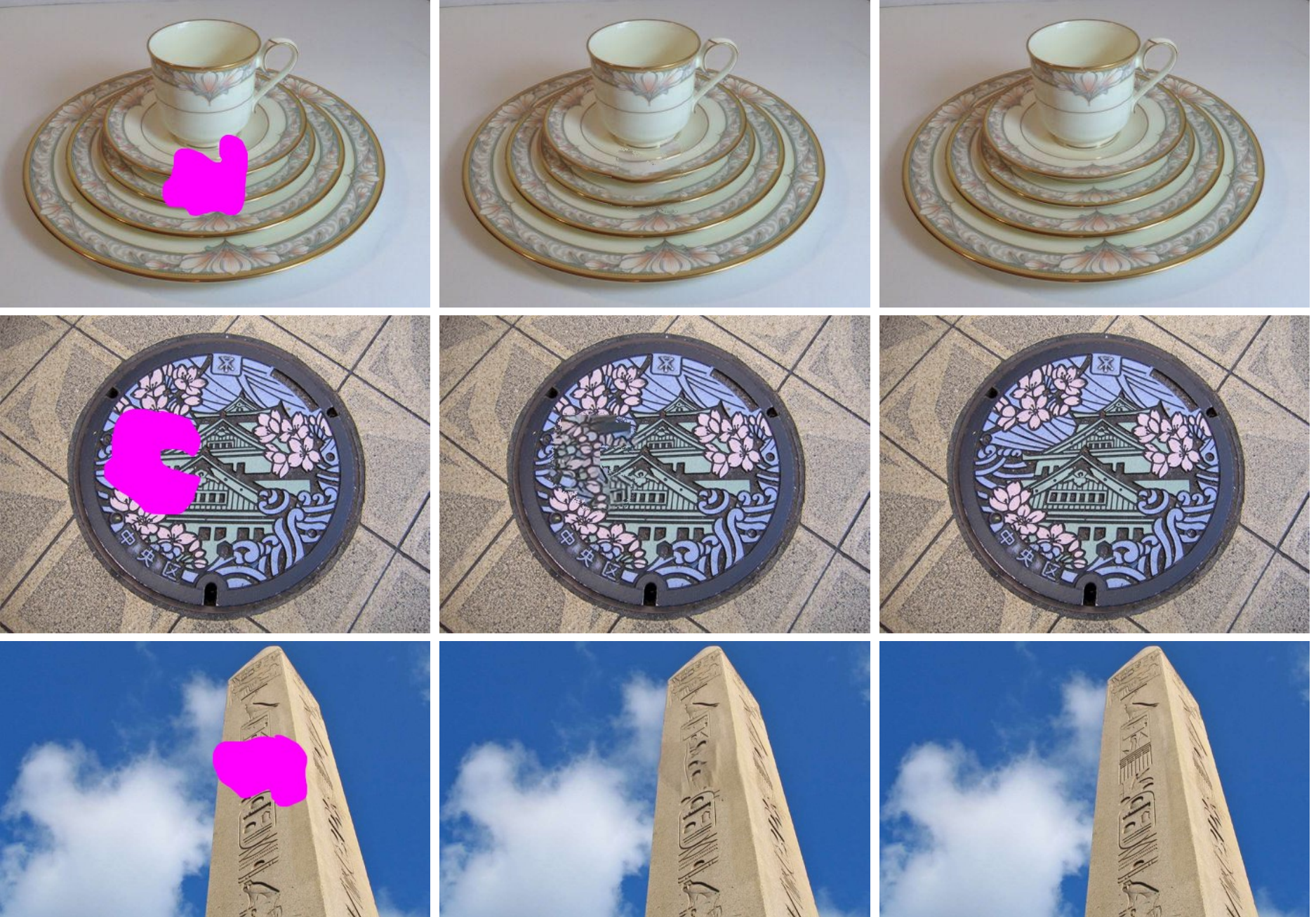}
\leftline{ \small \hspace{2.5cm} Input Image \hspace{4.0cm} CASI+ \hspace{4.0cm} Ground-Truth }
\caption{ Examples of Inpainting Results on Irregular Corrupted Regions }
\label{fig:irregular-mask}
\end{figure*}

\subsection{General and In-the-wild Case}
This section investigates how the proposed method perform on general and in-the-wild cases. The first experiment in this section is to test the proposed method on high-resolution real images that are collected out of ImageNet and Paris StreetView dataset. The qualitative results of the first experiment are shown in Figure~\ref{fig:real-images}. The resolution of the input images in Figure~\ref{fig:real-images} are $430 \times 645$, $708 \times 1062$ and $426 \times 570$. The results verify that our proposed method could perform well on in-the-wild cases.

The second experiment in this section is to test the proposed method on real images with irregular corrupted region. The qualitative results of the second experiment are displayed in Figure~\ref{fig:irregular-mask}. These input images are also collected in-the-wild out of ImageNet and Paris StreetView datasets and their resolutions are $357 \times 500$, $332 \times 450$ and $332 \times 450$ respectively. The results suggest that the proposed algorithm is capable of repairing images with irregular corrupted region.

\subsection{Investigation of Generalization Ability}
This section investigates the generalization ability of the CASI model. If the CASI model has weak generalization ability and overfits the training data, it may predict what it memorize from the training data. Thus we conduct a nearest neighbor inpainting (NN-inpainting) experiment. For each testing input image, we search for the most matching patch from the training dataset to complete the image, using the algorithm proposed in \cite{Hays:2007}. The qualitative results of NN-inpainting are displayed in Figure~\ref{fig:internal}. The CASI results (in Figure~\ref{fig:internal}d) are quite different from the NN-inpainting results and demonstrate the superiority in preserving both appearance and structure coherence, which indicates that the CASI model does not simply copy or memorize patch from the training dataset while repairing the input images.

\subsection{Comparison with the State of the Art}
We compare our proposed CASI model trained using the joint loss with other $4$ state-of-the-art image inpainting methods, including Content-Aware Fill~\cite{barnes2011patchmatch}, StructCompletion~\cite{huang2014image}, ImageMelding~\cite{darabi2012image} and Context Encoder~\cite{pathak2016context}. As shown in Figure~\ref{fig:more}, methods~\cite{barnes2011patchmatch},~\cite{huang2014image},~\cite{darabi2012image} without using neural network fail to recover the dog face in the first sample, extend the bus window in the second sample and connect the snake body in the third sample. These methods fail to recover the high-level semantics. Context Encoder struggles to display clear structure while the proposed CASI shows visually acceptable results in Figure~\ref{fig:more}.

\begin{table}[!t]
\renewcommand{\arraystretch}{1.3}
\caption{Quantitative results on Paris StreetView}
\label{table:paris}
\centering
\small
\begin{tabular}{l|c|c|c}
\hline
Method                      & mean $L1$ & mean $L2$ & PSNR\\
                            & error     & error     &\\
\hline\hline
PatchMatch                  & 12.59\%   &  3.14\%   &  16.82dB \\
NeuralPatch                 & 10.01\%   &  2.21\%   &  18.00dB \\
StructCompletion& 9.67\%   &  2.07\%   &  18.03dB \\
ImageMelding   &  9.55\%   &  2.19\%   &  18.05dB \\
Context Encoder             &  9.37\%   &  1.96\%   &  18.58dB \\
CASI                        &  \textbf{7.49\%}   & \textbf{1.37\%}   &  \textbf{20.37dB}\\
\hline
\end{tabular}

\end{table}

\begin{table}
\renewcommand{\arraystretch}{1.3}
\caption{Similarity Indices on ImageNet-20.}
\label{table:ssim}
\centering
\small
\begin{tabular}{l|c|c|c}
\hline
Method                     & SSIM  & FSIM & FSIMc\\
\hline\hline
Context Encoder            & 0.2579    &  0.6977   &  0.6899 \\
CASI,$L2$                  & 0.5196    &  0.6255   &  0.6202 \\
CASI,$L2$ + $per$          & 0.4927    &  0.6843   &  0.6779 \\
CASI,$L2$ + $adv$          & 0.5141    &  0.7202   &  0.7148 \\
CASI,$L2$+$adv$+$per$      & \textbf{0.5198}    &  \textbf{0.7239}   &  \textbf{0.7187} \\
\hline\hline
$\lambda_{per}=0$ & 0.5141 & 0.7202 & 0.7148\\
$\lambda_{per}=0.2$ & 0.5198 & 0.7239 & 0.7187\\
$\lambda_{per}=0.4$ & 0.5093 & 0.7203 & 0.7149\\
$\lambda_{per}=0.7$ & 0.4951 & 0.7163 & 0.7108\\
\hline
\end{tabular}
\end{table}

\begin{table}
\renewcommand{\arraystretch}{1.3}
\caption{Local Entropy Errors on ImageNet-20.}
\label{table:entropy}
\centering
\small
\begin{tabular}{l|c|c}
\hline
Method                      & LEMSE & LEMAE\\
\hline\hline
Context Encoder             &  0.5872   &  0.5391 \\
CASI,$L2$                  &   1.8926  &  1.0795 \\
CASI,$L2$ + $per$          &  0.8454   &  0.7219 \\
CASI,$L2$ + $adv$          &  0.4869   &  0.4945 \\
CASI,$L2$+$adv$+$per$      &  \textbf{0.4611}   &  \textbf{0.4847} \\
\hline\hline
$\lambda_{per}=0$ & 0.4869 & 0.4945 \\
$\lambda_{per}=0.2$ & 0.4611 & 0.4847 \\
$\lambda_{per}=0.4$ & 0.4470 & 0.4759 \\
$\lambda_{per}=0.7$ & 0.4492 & 0.4771 \\
\hline
\end{tabular}
\end{table}

The second experiment in this section compares our method with other state-of-the-art inpainting methods \cite{contentAwareFill, barnes2011patchmatch, yang2016high, huang2014image, darabi2012image, pathak2016context} on the Paris StreetView dataset. Table~\ref{table:paris} shows the quantitative results. Results from PatchMatch \cite{barnes2011patchmatch}, Neural Patch Synthesis (NeuralPatch) and Context Encoder are collected from \cite{yang2016high}, \cite{yang2016high} and \cite{pathak2016context}, respectively. As shown in Table~\ref{table:paris}, our results exceed others by a considerable margin under all three measures. Our method outperforms the second best by 1.58\% in mean $L1$ error, 0.53\% in mean $L2$ error and 1.56dB in PSNR.

\subsection{Investigation of Criteria for Inpainting}

\begin{table*}[!t]
\renewcommand{\arraystretch}{1.3}
\caption{Semantic Errors on ImageNet-20}
\label{table:SME}
\centering
\small
\begin{tabular}{l|c|c|c|c|c|c}
\hline
Method                      & SME-r50 & SME-r101 & SME-r152 & SME-r200 & SME-v16 & SME-v19\\
\hline\hline
baseline                    &  0.2063 & 0.1735 & 0.1852 & 0.2063 & 0.1794 & 0.2086\\
Context Encoder             &  0.1467 & 0.1462 & 0.1442 & 0.1467 & 0.1001 & 0.1123\\
CASI,$L2$                  &  0.1862 & 0.1908 & 0.1886 & 0.1877 & 0.1444 & 0.1652\\
CASI,$L2$ + $per$          &  0.1542 & 0.1631 & 0.1671 & 0.1626 & 0.1213 & 0.1384\\
CASI,$L2$ + $adv$          &  0.1276 & 0.1359 & 0.1349 & 0.1362 & 0.0846 & 0.0952\\
CASI,$L2$+$adv$+$per$      &  \textbf{0.1070} &  \textbf{0.1180} & \textbf{0.1201} & \textbf{0.1200} & \textbf{0.0721} & \textbf{0.0775}\\
\hline\hline
$\lambda_{per}=0$ & 0.1276 & 0.1360 & 0.1350 & 0.1363 & 0.0846 & 0.0952\\
$\lambda_{per}=0.2$ & 0.1070 & 0.1180 & 0.1201 & 0.1200 & 0.0721 & 0.0775\\
$\lambda_{per}=0.4$ & 0.1074 & 0.1125 & 0.1218 & 0.1215 & 0.0704 & 0.0767\\
$\lambda_{per}=0.7$ & 0.0994 & 0.1126 & 0.1117 & 0.1131 & 0.0632 & 0.0702\\
\hline
\end{tabular}
\end{table*}

In this section, we use more criteria to evaluate CASI and Context Encoder, and propose two new criteria for semantic inpainting. There are three major experiments. In the first experiment, we evaluate inpainting methods using structural similarity index (SSIM) \cite{wang2004image} and feature similarity index (FSIM) \cite{zhang2011fsim}. These indices are originally applied to image quality assessment (IQA) that attempts to quantify the visibility of differences between two images. Here we investigate the visual differences between inpainting results and their corresponding ground truth. Thus we test inpainting methods using the two IQA indices. SSIM is a classical index defined by structural similarity while FSIM is the state of the art based on two low-level features, phase congruency (PC) and gradient magnitude. FSIM is defined in Eq. (\ref{fsim}),
\begin{equation} \label{fsim}
FSIM = \frac{\sum S_{PC}(x) \cdot S_G(x) \cdot PC_m(x)}{\sum PC_m(x)},
\end{equation}
where $S_{PC}(x)$ and $S_G(x)$ are PC similarity and gradient similarity respectively at position $x$, and $PC_m(x)$ is the PC value of $x$ as a weight. As shown in Table~\ref{table:ssim}, all CASI models achieve higher similarity with the ground truth than Context Encoder under SSIM, FSIM and FSIMc (FSIM for color image). It indicates that our method not only recovers more consistent structures but also synthesizes content with higher visual quality. However, SSIM and FSIM are still biased towards blurry results of CASI, $L2$ ($+l_{per}$).

In the second experiment, we introduce a novel local entropy error to rate blurry predictions more accurately. Entropy in texture analysis is a statistic characterizing the texture within an image region, as defined in \cite{rafael2003digital}. The local entropy at a pixel is defined as the entropy within a $9\times9$ neighborhood of the pixel. We define local entropy error as the mean squared error (denoted as LEMSE) or the mean absolute error (LEMAE) of local entropy within the synthesized region. As shown in Table~\ref{table:entropy}, our proposed CASI delivers the lowest LEMSE and LEMAE among all methods. In addition, CASI with $L2$ loss and CASI with $L2 + per$ loss achieve the largest and second largest errors under both LEMSE and LEMAE, which is consistent with most of the visual results (a subset is given in Figure~\ref{fig:internal}) and confirms that our proposed local entropy error is capable of rating over-smooth results accurately.

In the third experiment, we propose a high-level criterion, semantic error, which aims at measuring how successful an inpainting method recovers the semantics. Semantic error (SME) is defined with respect to a pre-trained image classifier that outputs a probability of the image being part of each possible category. SME is based on two probabilities that the groundtruth image and the synthesized image belong to the groundtruth category respectively. It is formulated as in the following equation,
\begin{equation} \label{SME}
SME = \frac{1}{n} \sum_{i=1}^n max(0,P_{x_{i}}^{y_{i}} - P_{z_{i}}^{y_{i}}),
\end{equation}
where $n$ is the number of testing samples, $x_{i}$, $z_{i}$ and $y_{i}$ are the groundtruth image, synthesized image (with real context) and the groundtruth category of the $i$-th sample. $P_{x_{i}}^{y_{i}}$ is the probability that image $x_{i}$ belongs to category $y_{i}$, estimated by a pre-trained classifier (e.g., residual network \cite{he2016deep} or VGG network\cite{Simonyan14c}). Here we associate the probability of assigning the correct label with our semantic error because we focus on to what extent a corruption ``makes a dog unlike a dog'' and to what extent the restored content ``makes a dog look like a dog again''. A baseline model simply fills the missing region with the mean pixel value. The SME of this baseline measures how much a corrupted region harms the semantic information of an image. In Table~\ref{table:SME}, SME-r$L$ represents the SME achieved by applying an $L$-layer residual network as the classifier while SME-v$L$ represents the SME achieved by adopting an $L$-layer VGG network as the classifier. Notice that our feature network is simpler than the ResNets used for estimating SME, which implies that harvesting knowledge using a low-capacity model can reduce the SME estimated by a high-capacity classifier. As shown in Table~\ref{table:SME} shows, our proposed network outperforms other inpainting methods by achieving the smallest semantic error.

Perceptual loss weight is also investigated on the above new criteria for semantic inpainting, as shown in the lower part of Table~\ref{table:ssim}, Table~\ref{table:entropy} and Table~\ref{table:SME}. $\lambda_{per}=0.7$ performs better on similarity indices and semantic errors while $\lambda_{per}=0.4$ demonstrates better results on local entropy error. To compromise different criteria, $\lambda$ is chosen from $0.2$ to $0.4$.

\section{Conclusion}
In this paper, we have presented a fully convolutional generative adversarial network with a context-aware loss function for semantic inpainting. This network employs a fully convolutional architecture in the generator, which does not have a fully connected layer as the bottleneck layer. The joint loss includes a perceptual loss to capture semantic information around the synthesized region. In addition, we have developed two new measures for evaluating sharpness and semantic validity respectively. In summary, our method delivers state-of-the-art results in qualitative comparisons and under a wide range of quantitative criteria.


%



\ifCLASSOPTIONcaptionsoff
  \newpage
\fi
\newpage
\IEEEtriggeratref{-1}
\bibliographystyle{IEEEtran}
\bibliography{egbib}
\end{document}